\documentclass[10pt,twocolumn,letterpaper]{article}

\usepackage{iccv}
\usepackage{times}
\usepackage{epsfig}
\usepackage{graphicx}
\usepackage{amsmath}
\usepackage{amssymb}
\usepackage{booktabs}
\usepackage{algorithm}
\usepackage{algorithmic}
\usepackage{multirow}
\usepackage{subcaption}
\usepackage[accsupp]{axessibility}  

\usepackage[breaklinks=true,bookmarks=false]{hyperref}

\usepackage[capitalize]{cleveref}
\crefname{section}{Sec.}{Secs.}
\Crefname{section}{Section}{Sections}
\Crefname{table}{Table}{Tables}
\crefname{table}{Tab.}{Tabs.}
\crefname{figure}{Fig}{Figs.}

\iccvfinalcopy 

\def\iccvPaperID{4127} 

\ificcvfinal\fi

\begin{document}

\title{Backpropagation Path Search On Adversarial Transferability}

\author{Zhuoer Xu$^{1}$, \ 
Zhangxuan Gu$^{1}$,
Jianping Zhang$^{2}$\thanks{Corresponding author.} \ ,
Shiwen Cui$^{1}$,
Changhua Meng$^{1}$,
Weiqiang Wang$^{1}$, \\ 
$^{1}$Tiansuan Lab, Ant Group \\
$^{2}$Department of Computer Science and Engineering, The Chinese University of Hong Kong \\
{\tt \small \{xuzhuoer.xze,guzhangxuan.gzx,donn.csw,changhua.mch,weiqiang.wwq\}@antgroup.com}\\
{\tt \small jpzhang@cse.cuhk.edu.hk}
}

\def\pas{PAS}
\def\subsetone{Subset1000}
\def\subsetfive{Subset5000}
\def\skipconv{SkipConv}
\def\relu{ReLU}
\def\skiprelu{Lin\relu{}}
\def\skipgrad{SkipGrad}
\def\x{\boldsymbol{x}}
\def\sign{\text{sign}}
\newcommand{\xadv}[1]{\x{}_{adv}^{#1}}
\newcommand{\z}[1]{\boldsymbol{z}_{#1}}
\newcommand{\ens}[1]{$_{\text{ens{#1}}}$}

\maketitle
\ificcvfinal\thispagestyle{empty}\fi

\begin{abstract}

Deep neural networks are vulnerable to adversarial examples, dictating the imperativeness to test the model's robustness before deployment.
Transfer-based attackers craft adversarial examples against surrogate models and transfer them to victim models deployed in the black-box situation.
To enhance the adversarial transferability, structure-based attackers adjust the backpropagation path to avoid the attack from overfitting the surrogate model.
However, existing structure-based attackers fail to explore the convolution module in CNNs and modify the backpropagation graph heuristically, leading to limited effectiveness.
In this paper, we propose backPropagation pAth Search (\pas{}), solving the aforementioned two problems.
We first propose \skipconv{} to adjust the backpropagation path of convolution by structural reparameterization.
To overcome the drawback of heuristically designed backpropagation paths, we further construct a Directed Acyclic Graph (DAG) search space, utilize one-step approximation for path evaluation and employ Bayesian Optimization to search for the optimal path.
We conduct comprehensive experiments in a wide range of transfer settings, showing that \pas{} improves the attack success rate by a huge margin for both normally trained and defense models.
\end{abstract}

\section{Introduction}

\begin{figure}[t]
\begin{subfigure}{1.0\linewidth}
\includegraphics[width=\textwidth]{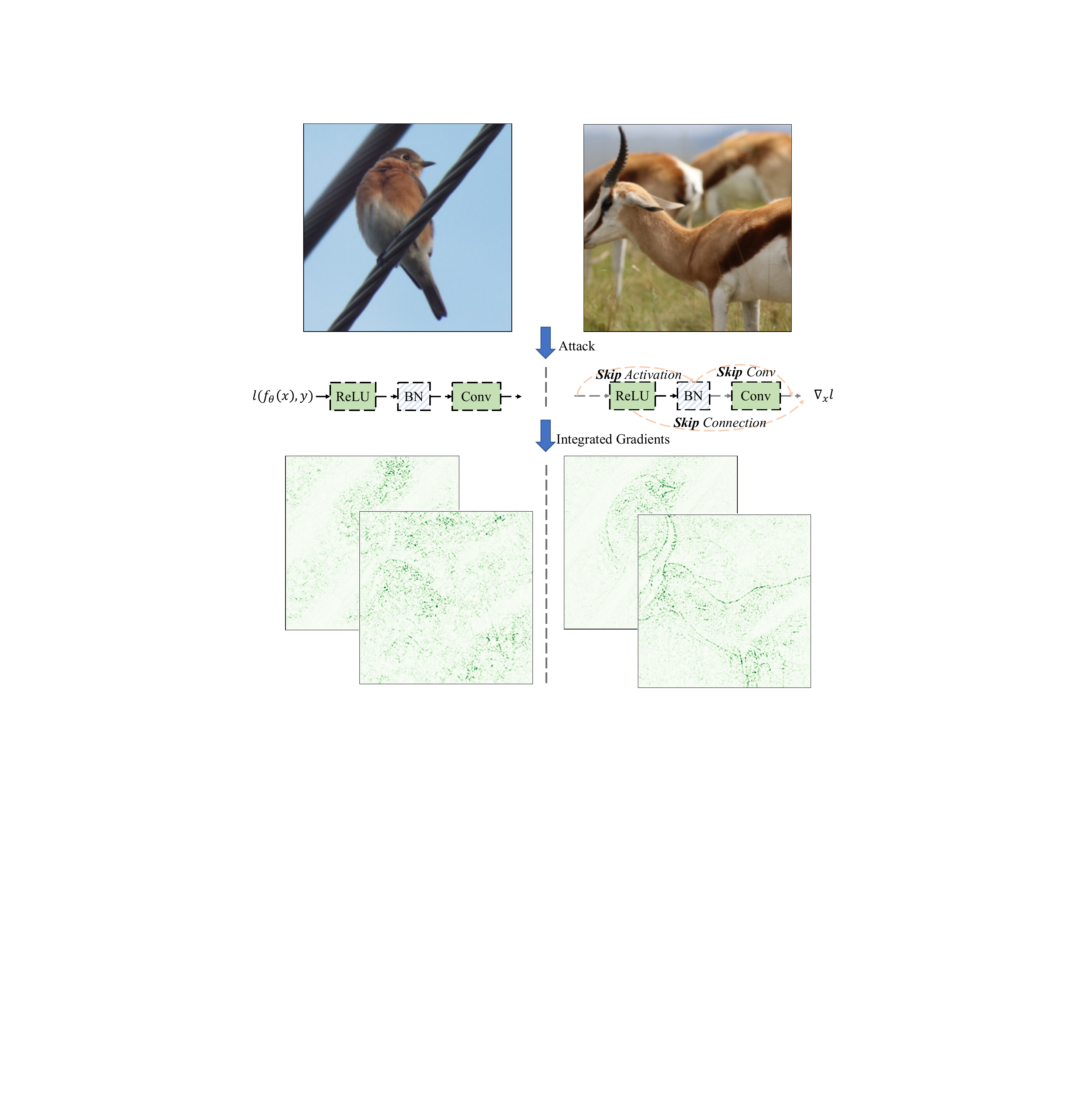}
\end{subfigure}
\begin{subfigure}{0.55\linewidth}
\centering
\caption{Original Surrogate}
\end{subfigure}
\begin{subfigure}{0.35\linewidth}
\centering
\caption{PAS Surrogate}
\end{subfigure}
\caption{
Feature attribution with Integrated Gradients for both the original surrogate and the surrogate with the searched backpropagation path (\ie, \pas{}).
%
PAS explicitly enhances the object attribution for classification and reduces the overfitting of the surrogate model to the irrelevant background, which demonstrates the effectiveness and interpretability of searched backpropagation path.
}
\label{fig:ig-pas}
\end{figure}

Deep neural networks (DNNs) are vulnerable to adversarial examples~\cite{szegedy2013adv} despite their success in a wide variety of applications \cite{he2016resnet,guo2017deepfm,kenton2019bert,gu2022diffusioninst}.
It is imperative to devise effective attackers to identify the deficiencies of DNNs beforehand, which serves as the first step to improving the model's robustness.
White-box attackers~\cite{madry2018pgd,Carlini017cw,croce2020autoattack,zhang2023zjp3diffusion} have complete access to the structures and parameters of victim models and effectively mislead them.
However, DNNs are generally deployed in the black-box situation.
To this end, transfer-based attackers, as typical black-box attackers form without access to information about victim models, have drawn increasing attention in the research community~\cite{LiuCLS17ensi,xie2019di,zhang2023atk1}.

It is widely known that adversarial examples, crafted following a white-box situation against a surrogate model, are transferable to the black-box victim models due to the linear nature of DNNs~\cite{fgsm}.
To boost adversarial transferability, various methods have been proposed on different aspects, \eg,
momentum terms~\cite{dong2018mi,lin2019nisi},
data augmentation~\cite{xie2019di,DongPSZ19ti},
structure augmentation~\cite{wu2019sgm,guo2020linbp,li2020ghost,fang2022llta},
ensemble \cite{LiuCLS17ensi,xiong2022svre},
and intermediate features \cite{AdityaGaneshan2019fda,zhang2022naa}.
The common characteristic of the above attackers is that they reduce the overfitting of the attack on the surrogate model.

In this paper, we focus on structure-based attackers~\cite{wu2019sgm,guo2020linbp,li2020ghost,fang2022llta}, which directly rectify the backpropagation path to alleviate the overfitting issue and expose more transferability of adversarial attacks.
For example, SGM \cite{wu2019sgm} and LinBP \cite{guo2020linbp} reduces the gradient from residual and nonlinear activation modules, respectively.
However, existing structure-based attackers suffer from two critical problems circumventing their transferability. 
(1) They neglect the convolution module, which plays a significant role in extracting features as a basic but vital module in CNNs.
The lack of adjustment for convolution in backpropagation prevents the exploitation of gradients from critical features and leads to limited effectiveness.
(2) Their modification of the backpropagation path follows a heuristic manner by predefined hyper-parameters, so the selected path is non-optimal.

To explore the backpropagation of the convolution module, we follow SGM~\cite{wu2019sgm} to explore the backpropagation path with skip connections.
Note that the inherent structure of convolution does not have skip connections for adjustment.
Thus, we propose \skipconv{}, which decomposes the original convolution kernel into one skip kernel acting as a skip connection and the corresponding residual convolution kernel.
With the two decomposed kernels, \skipconv{} calculates forward as usual but it is convenient to modify the backpropagation gradient via the skip kernel.

Meanwhile, we endeavor to not only resolve the heuristic problem but also unify existing structure-based attackers.
Especially, we analogize the structure-based adversarial attack as a transferable backpropagation path search problem.
We propose a unified and flexible framework for backPropagation pAth Search, which consists of search space, search algorithm, and evaluation metric, namely \pas{}.
Intending to explore transferable backpropagation paths, we construct a DAG combining the skip paths of convolution, activation, and residual modules in DNNs as the search space.
Next, we employ Bayesian Optimization to search for the optimal path and avoid heuristic designs.
To reduce the additional overhead introduced by such a black-box search, we adopt a one-step approximation schema to efficiently evaluate the paths.
Extensive experiments on the subsets of ImageNet from different surrogate models demonstrate the effectiveness of \pas{} against both normally trained and defense models in comparison with the baseline and state-of-the-art (SOTA) attackers.

Our main contributions can be summarized as follows:
\begin{itemize}
    \item We propose \skipconv{}, which decomposes a convolution kernel into one skip kernel and the residual kernel via structural reparameterization. Such decomposition is convenient for the exploration of the convolution module during backpropagation for boosting adversarial transferability.
    \item We analogize the structure-based adversarial attack as a transferable backpropagation path search problem. Thus, we propose a unified framework \pas{} for backpropagation path search. \pas{} employs Bayesian Optimization to search for transferable paths in DAG-based search space. The search overhead is further reduced by one-step approximation evaluation.
    \item We conduct comprehensive experiments in a wide range of transfer settings.
    \pas{} greatly improves the attack success rate for normally trained models in all cases and achieves a huge margin of 6.9\%$\sim$24.3\% improvement against defense models.
    The results demonstrate the generality of \pas{} with various surrogate models on two benchmarks.
\end{itemize}

\begin{figure*}[h]
\centering
\includegraphics[width=0.9\textwidth]{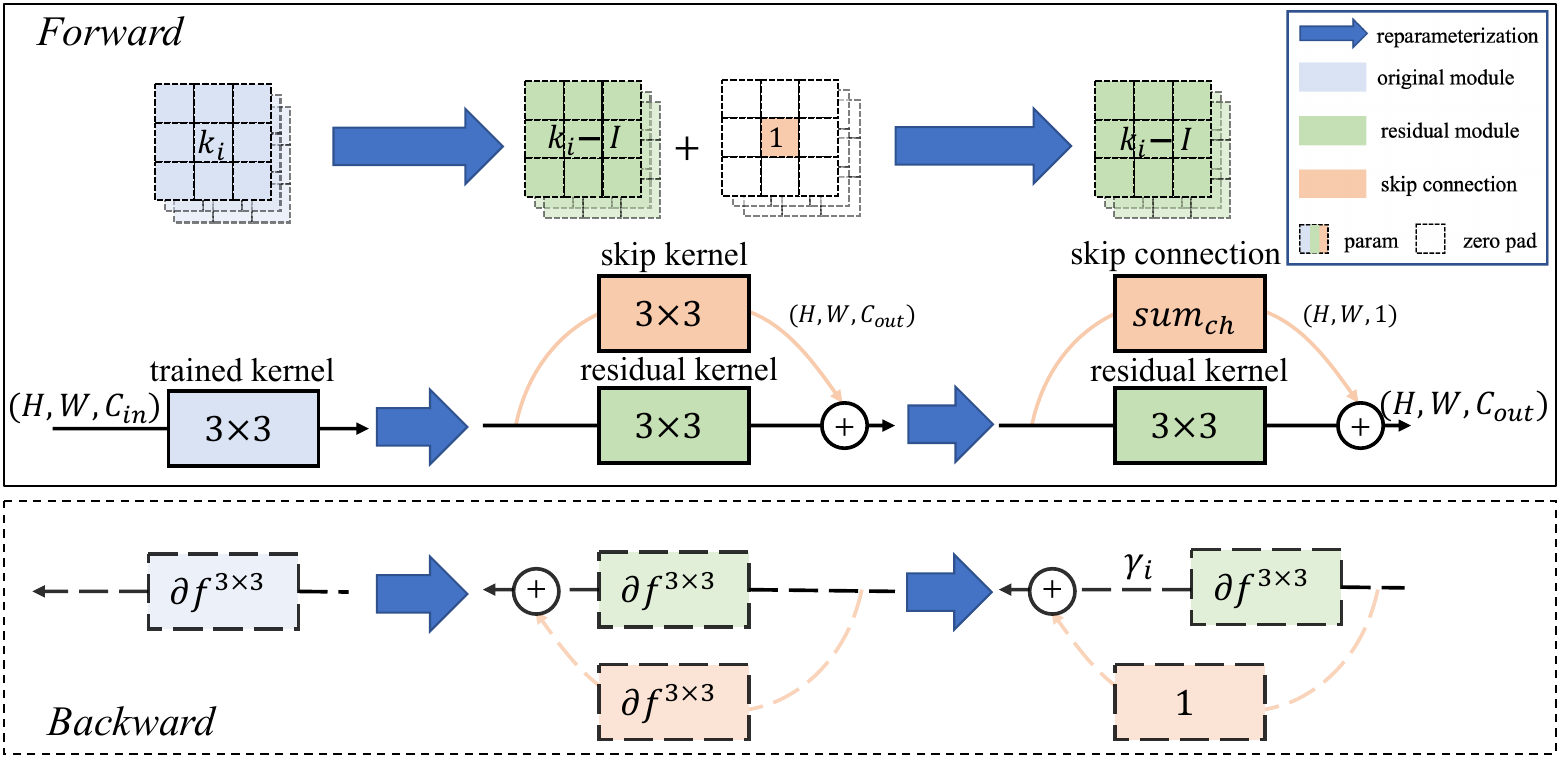}
\caption{
\skipconv{}: structural reparameterization of the convolution module.
Take the $3\times3$ convolution as an example.
According to convolution distributivity, the original kernel $k_i$ is decomposed into the sum of the skip kernel $I$ and the corresponding residual kernel $k_i-I$.
The skip connection is implemented via structural reparameterization.
\skipconv{} calculates forward as the original convolution but backpropagates the loss in a skip connection form.
$\gamma_i$ is introduced to control the gradient from the residual kernel.
}
\label{fig:skipconv}
\end{figure*}

\section{Preliminary}

Given a clean example $\x{}$ with class label $y$ and a victim model $f_{\theta}$ parameterized by $\theta$,
the goal of an adversary is to find an adversarial example $\xadv{}$,
which is constrained by $L_p$ norm with a bound $\epsilon$,
to fool the model into making an incorrect prediction:
\begin{equation}
\label{eq:adv}
    f_{\theta}(\xadv{}) \not = y, \text{ where }\lVert \xadv{} - \x{} \rVert_{p} \leq \epsilon
\end{equation}

In the white-box situation, FGSM~\cite{fgsm} perturbs the clean example $\x{}$ for one step by the amount of $\epsilon$ along the gradient direction.
As an iterative version, I-FGSM~\cite{kurakin2018ifgsm} perturbs $\x{}$ for $T$ steps with smaller step size $\eta$ and achieves a high attack success rate:
\begin{equation}
    \xadv{t+1} = \Pi_{\epsilon}^{\x{}} \left(
        \xadv{t} +
        \eta  \cdot \sign{}\left(
            \nabla_{\x{}} l\left(
                f_{\theta}\left( \xadv{t} \right), y
            \right)
        \right)
    \right)
\end{equation}

In the absence of access to the victim model $f_{\theta}$, transfer-based attackers craft adversarial examples against a white-box surrogate model $f_{\theta_s, \Gamma}$ with structure hyper-parameters $\Gamma$ (\eg, hyper-parameters for residual and activation modules) to achieve \cref{eq:adv}:
\begin{equation}
\begin{aligned}
    &\xadv{t+1} = \Pi_{\epsilon}^{\x{}} \left(
        \xadv{t} +
        \eta  \cdot \sign{}\left(
            \nabla_{\x{}} l\left(
                f_{\theta_s, \Gamma}\left( \xadv{t} \right), y
            \right)
        \right)
    \right)
\end{aligned}
\end{equation}

Backpropagation is essential in the process of adversarial example generation.
Classical DNNs consist of several layers, \ie, $f =  f_1 \circ \cdots \circ f_L$, where $i \in \{ 1, \ldots L \}$ is the layer index, and $\z{i} = f_{i} (\z{i-1})$ indicates the intermediate output and $\z{0}=\x{}$.
According to the chain rule in calculus, the gradient of the loss $l$ \wrt input $\x{}$ can be then decomposed as:
\begin{equation}
    \frac{\partial l}{\partial \x{}} =
    \frac{\partial l}{\partial \z{L}}
    \frac{\partial f_{L}}{\partial \z{L-1}}
    \cdots
    \frac{\partial f_{1}}{\partial \z{0}}
    \frac{\partial \z{0}}{\partial \x{}}
\end{equation}
In this case, a single path is used for the gradient propagation backward from the loss to the input.
Extending $f$ to a ResNet-like (with skip connections) network, the residual module in layer $i$ where $f^{res}_{i}(\z{i-1}) = \z{i-1} + f_{i}(\z{i-1})$ decomposes the gradient as:
\begin{equation}
\begin{aligned}
    \frac{\partial l}{\partial \z{0}} &=
    \frac{\partial l}{\partial \z{L}}\cdots
    \frac{\partial f^{res}_{i}}{\partial \z{i-1}} \cdots
    \frac{\partial \z{0}}{\partial x} \\
    &=
    \frac{\partial l}{\partial \z{L}}\cdots
    \left( 1 + \frac{\partial f_{i}}{\partial \z{i-1}} \right)\cdots
    \frac{\partial \z{0}}{\partial x}
\end{aligned}
\end{equation}
The residual module provides a gradient highway as mentioned in \cite{he2016resnet}.

\section{Methodology}

In this section, we first introduce how to expand the backpropagation path of convolution via structural reparameterization in \cref{sec:skipconv}.
Then, we demonstrate the three parts of PAS (\ie, search space, search algorithm, and evaluation metric) in \cref{sec:pas}.
Finally, we present the overall process.
%
%

\begin{figure*}[ht]
\centering
\includegraphics[width=0.95\textwidth]{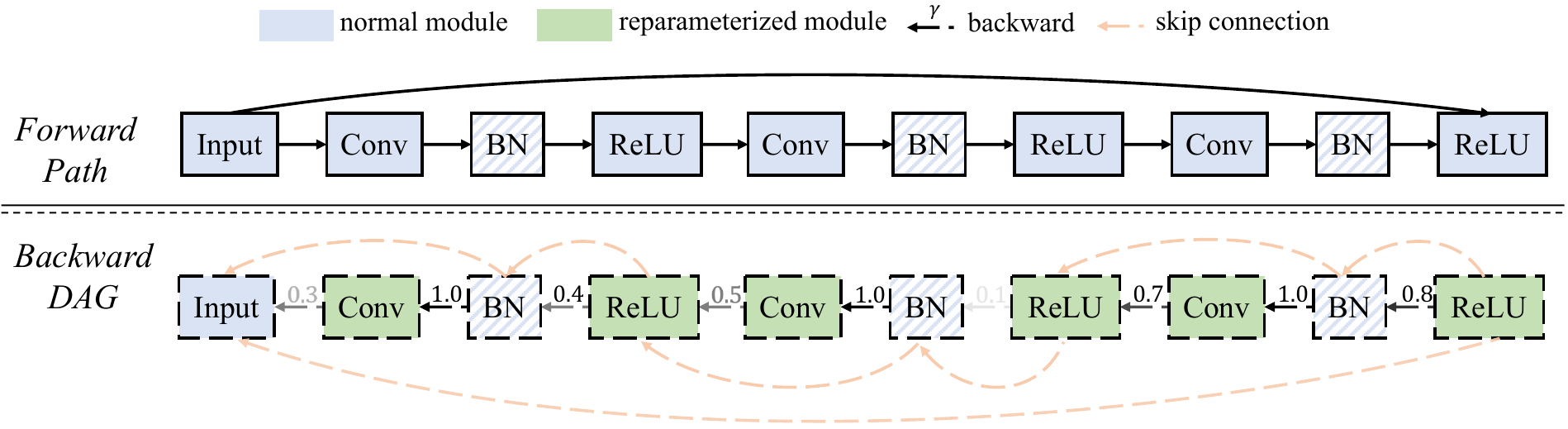}
\caption{
Example of backpropagation DAG. We construct the DAG by combining all the backpropagation paths of reparameterized modules. The color transparency indicates the weight $\gamma$ of the corresponding path to control its weight.
}
\label{fig:dag}
\end{figure*}

\subsection{Skip Convolution}
\label{sec:skipconv}

Skip connections in backpropagation allow easier generation of highly transferable adversarial examples \cite{wu2019sgm}.
However, as a basic module to extract diverse features, convolution is neglected for adversarial transferability.
Existing structure-based attackers lack the exploitation of gradients from critical features shared among different DNNs.
We propose \skipconv{} to fill in the missing piece of the puzzle.
The key of \skipconv{} is not to decompose the convolution in the backpropagation path without affecting the forward pass.

To achieve the characteristic, we reparameterize the structure of convolution $f_i^{conv}$ with kernel $k_i$ as shown in the \cref{fig:skipconv}.
Specifically, we decompose the original convolution kernel $k_i$ into the sum of two kernels, \ie, $k_i = k_i^{res} + k_i^{skip}$.
According to the distributivity of convolution, the decomposed kernels calculate forward as usual, \ie, $f_{i}^{conv}(\z{i-1};k_i)  = f_{i}^{conv}(\z{i-1};k_i^{res}) + f_{i}^{conv}(\z{i-1};k_i^{skip})$.
Next, we view the skip convolution as an identity to make the decomposed convolution backpropagate loss in a skip connection, \ie, $\partial f_{i}^{conv}(\z{i-1};k_i^{skip}) / \partial \z{i-1} = \z{i-1}$.
Inspired by RepVGG \cite{ding2021repvgg}, it is implemented by constructing a $1\times 1$ skip kernel $I$ with all values 1 and zero-padding $I$ to the shape of $k_i$, \ie, $f_{i}^{conv}(\z{i-1};I)=sum_{ch}(\z{i-1})$.
%
%
%

All in all, we decompose the kernel as the sum of a constant skip kernel $I$ and the corresponding residual kernel $k_i-I$.
The skip kernel plays the same role as the skip connection, only calculating the sum of each channel to change the channel dimension.
In this way, we reparameterize the structure of convolution into the skip connection and the residual convolution:
%
%
\begin{equation}
\begin{aligned}
    f_{i}^{conv} (\z{i-1}; k_i)
    = sum_{ch}(\z{i-1}) \\
    + \gamma_i \cdot f_{i}^{conv}(\z{i-1}; k_i - I) + C
\end{aligned}
\end{equation}
where the decay factor $\gamma_i \in [0,1]$ is introduced as the weight of the residual gradient and $C$ is equal to $(1-\gamma_i) \cdot f_{i}^{conv}(\z{i-1}; k_i - I)$ without gradient backward.
Such \skipconv{} requires no fine-tuning since it calculates forward as usual.
For backpropagation, $\gamma_i$ is used to relatively adjust the gradient, \ie,
$1 + \gamma_i \cdot \partial f_{i}^{conv}(\z{i-1};k_i - I) / \partial \z{i-1}$.

\subsection{PAS: Backpropagation Path Search}
\label{sec:pas}

In this part, we introduce how \pas{} searches for highly transferable paths, which are evaluated by one-step approximation, in the DAG-based space.

\subsubsection{Backpropagation DAG}
\label{sec:dag}

Unlike works that use the existing backpropagation path of the surrogate model (\eg, residual module), \pas{} reparameterizes original modules with skip connections and expands the path as a DAG via structural reparameterization.

\noindent \textbf{Skip Activation.}
\relu{} is a common activation module in neural networks.
\cite{guo2020linbp} demonstrates that the gradient of \relu{} is sparse, which degrades adversarial transferability.
The gradient of ReLU is propagated backward as $\partial f^{\relu{}}_{i}/\partial \z{i-1} = W_i M_i W_{i + 1}$, where $M_i$ is a diagonal matrix whose entries are $1$ if the corresponding entris of $W_i^T \z{i-1}$ are positive and 0 otherwise.
LinBP~\cite{guo2020linbp} skips the \relu{} module and renormalizes the gradient passing backward as $\partial f^{ReLU}_{i}/\partial \z{i-1} = \alpha_i \cdot W_i W_{i - 1}$ where
$\alpha_i=\lVert W_i M_i W_{i - 1} \rVert_2 / \lVert W_i W_{i - 1} \rVert_2$.
However, the scalar $\alpha_i$ used for normalization needs to be calculated based on the weight of the front and back layers.
We further devise an approximation for $\alpha_i$ and reparameterize \relu{} as follows:
\begin{equation}
\begin{aligned}
    f_i^{\relu{}} (\z{i-1} ) =
    \hat{\alpha_i} \cdot \left(\z{i-1}
    + \relu{} \left( -\z{i-1} \right) \right) \\
    + (1 - \hat{\alpha_i}) \cdot \relu{}(\z{i-1})
\end{aligned}
\end{equation}
where $\hat{\alpha_i} = \lVert M_i \rVert_2 / \lVert z_{i-1} \rVert_2$ uses the sparsity as the estimation of the re-normalizing factor.

\noindent \textbf{Skip Gradient.}
SGM~\cite{wu2019sgm} introduces a decay parameter to control gradients from the existing skip connections, \ie, $ \partial f^{res}_{i} / \partial \z{i-1} = 1 + \gamma \cdot \partial f_{i} / \partial \z{i-1}$.
We use the same \skipgrad{} implemented by structural reparameterization:
\begin{equation}
    f_{i}^{res} (\z{i-1}) = \z{i-1} + \gamma_i \cdot f_{i}(\z{i-1}) + C
\end{equation}
where $C = (1-\gamma_i) \cdot f_{i}(\z{i-1})$ without gradient backward.

\noindent \textbf{Directed Acyclic Graph.}
We reparameterize the structure of diverse modules in DNNs and control the weight of paths by $\gamma$.
For each module's backpropagation path, we control the gradient backward via \skipconv{} and \skiprelu{}.
For cross-module paths, we use the existing skip connection as a highway for adversarial transferability.
By combining all the paths of the above modules, we construct the Directed Acyclic Graph (DAG) for gradient propagation backward.
As shown in \cref{fig:dag}, we use $\Gamma = \{ \gamma_i \}$ to control the weight of the residual path, and hence black-box optimization can be used to search for the most transferable paths.

\subsubsection{One-Step Approximation for Path Evaluation.}
\label{sec:approx}
To guide the search on the DAG, we evaluate the searched paths and further propose the one-step approximation to alleviate the large overhead consumed in the search process.
Intuitively, the highly transferable paths have a high attack success rate on all data for any victim model.
Thus, we evaluate the impact of different steps and samples for path evaluation, which will be detailed in the appendix.
The results experimentally verify that the one-step attack success rate of samples on only one white-box validation model is sufficient to distinguish paths' transferability.
Based on this, an approximate schema is adopted, \ie, we use such one-step evaluation as the estimation of transferability:
\begin{equation}
\begin{aligned}
   & s(\Gamma; \theta_s, \theta_m, N) =
    \frac{1}{N} \sum_{i=0}^N \mathbf{1} \{
    f_{\theta_m} (
         \\
           &\underbrace{
           \x{}^{(i)} +
            \epsilon  \cdot \sign{}(
                \nabla_{\x{}^{(i)}} l(
                    f_{\theta_s, \Gamma}( \x{}^{(i)} ), y^{(i)}
                )
            }_{\text{one-step adversarial examples crafted against path $\Gamma$}}\\
         &)
    )
    \not = y^{(i)}
    \} 
\end{aligned}
\label{eq:obj}
\end{equation}
Related techniques have been used in neural architecture search for evaluation \cite{darts}, gradient-based hyperparameter tuning \cite{JelenaLuketina2016ScalableGT} and fast adversarial training for attacks \cite{jia2022fat,xzr2022a2}.

\subsubsection{Unified Framework}

\begin{figure}[ht]
\centering
   \includegraphics[width=1.0\linewidth]{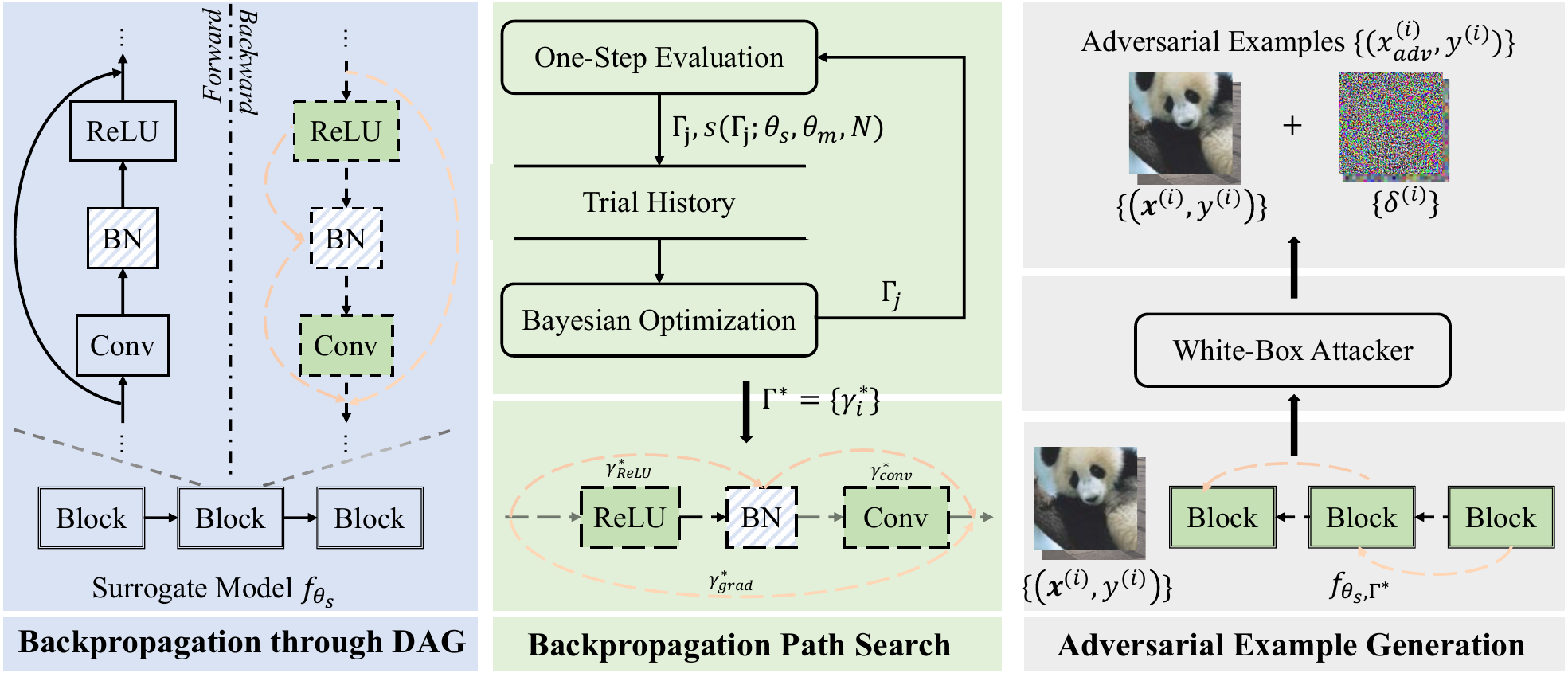}
\caption{
Overview of PAS.
\textbf{Left Block} demonstrates that PAS keeps the forward path and reparameterizes the modules in the backward pass.
Structural reparameterization expands the original graph into a DAG with reparameterized skip connections (dash lines).
\textbf{Middle Block} employs search algorithms (e.g., Bayesian Optimization) to find backpropagation paths based on their transferability evaluation.
\textbf{Right Block} illustrates that adversarial examples are crafted via the searched path of PAS surrogate.
}
\label{fig:pas-all}
\end{figure}

To optimize the above objective \cref{eq:obj}, we use Bayesian Optimization\footnote{https://optuna.org/} to search the structure parameters $\Gamma$ and combine it with Hyperband~\cite{li2017hyperband} to allocate resources for each trial of the sampled path.
The overall procedure is shown in \cref{alg:pas}.
We first search for the most transferable path $\Gamma^{*}$ of the surrogate model according to \cref{eq:obj} and then craft adversarial examples, which are transferred to unaccessible victim models.

In the search process, \pas{} reparameterizes the structure of the surrogate model and initializes $\Gamma$.
Bayesian Optimization is used to sample the backpropagation path $\Gamma_{k}$.
According to the sampled paths, adversarial examples for the validation dataset are crafted against $f_{\theta_s, \Gamma_k}$.
\pas{} calculates the attack success rate on the validation model and uses it as the feedback for Bayesian Optimization for the next iteration.
When predefined resources are exhausted, \pas{} uses the optimal structure $\Gamma^{*}$ to craft adversarial examples on the test set and transfers them to all victim models.

\begin{algorithm}
\caption{\pas{}: Backpropagation Path Search on Adversarial Transferability}
\label{alg:pas}
\textbf{Input}: Surrogate model $f_{\theta_s}$, validation model $f_{\theta_m}$, perturbation bound $\epsilon$, the number of attack steps $T$, the number of trials $N_{t}$

\begin{algorithmic}[1] 
\STATE Reparameterize the structure of $f_{\theta_s}$ as $f_{\theta_s, \Gamma}$

\FOR{$j$ $=$ $1, \ldots ,N_{t}$}
    \STATE sample $\Gamma_j$ by Bayesian Optimization according to the trial history
    \STATE evaluate $\Gamma_{j}$ by \cref{eq:obj} and add it to the history
\ENDFOR

\STATE select the most transferable path $\Gamma^{*}$ according to the history

\RETURN adversarial examples crafted against $f_{\theta_s, \Gamma^*}$

\end{algorithmic}
\end{algorithm}

\noindent \textbf{Flexibility.}
As a unified framework, \pas{} consists of three parts for extensions.
More effective and efficient searches can be achieved through different search algorithms and evaluation metrics.
It is flexible to use the new proposed augmentation in the backpropagation path to improve the diversity of the DAG-based search space.
For example, we explore structural reparameterization for the Transformer’s modules and use PAS to improve adversarial transferability between CNNs and Transformers in future work.

\begin{table*}[h]
\centering
\scalebox{0.95}{
\centering
\begin{tabular}{@{}llrrrrrrr|rrr@{}}
\toprule
                       & Attacker       & RN152          & DN201          & SE154          & IncV3          & IncV4          & IncRes    & ViT      & IncV3\ens{3}          & IncV3\ens{4}          & IncRes\ens{3}        \\ \midrule
\multirow{6}{*}{\rotatebox{90}{RN152}} & I-FGSM          & 99.91          & 51.00          & 26.32          & 23.50          & 22.58          & 18.72     &5.10     & 12.20          & 10.80          & 5.70     \\
& Ens & 99.94 & 56.68 & 38.64 & 27.56 & 30.68 & 24.06 & 6.16 & 13.22 & 10.90 & 6.92\\
& SVRE & 99.26 & 70.54 & 49.16 & 43.46 & 40.86 & 29.70 & 9.88 & 20.82 & 19.84 & 12.08\\
                      & MI           & 99.82          & 75.79          & 53.00          & 46.50          & 43.32          & 33.08     &15.28     & 24.20          & 22.04          & 16.10          \\
                      & DI           & 99.78          & 77.81          & 57.49          & 50.28          & 47.16          & 35.10    &10.40      & 35.97          & 32.81          & 20.16          \\
                       & SGM          & 99.87          & 82.76          & 61.90          & 53.16          & 49.24          & 43.30   & 11.72       & 31.57          & 27.77          & 20.84          \\
                       & IAA          & 99.87          & 95.06          & 82.46          & 76.34          & 71.04          & 58.34   & /       & 43.28          & 37.88          & 26.78          \\
                       & \pas{} & \textbf{99.96} & \textbf{96.76} & \textbf{84.98} & \textbf{83.82} & \textbf{78.82} & \textbf{77.18} & \textbf{50.26} & \textbf{59.34} & \textbf{54.46} & \textbf{44.74} \\
\midrule
\multirow{6}{*}{\rotatebox{90}{DN201}} & I-FGSM          & 59.08          & 99.89          & 40.60          & 33.80          & 32.46          & 23.80   & 6.54       & 18.16          & 15.30          & 10.40          \\
& Ens & 60.46 & 99.96 & 44.02 & 33.16 & 37.34 & 27.94 & 8.08 & 20.48 & 17.78 & 11.48 \\
& SVRE & 71.50 & 99.66 & 56.50 & 46.74 & 49.16 & 32.86 & 12.50 & 25.58 & 22.24 & 16.26 \\
                      & MI           & 76.39          & 99.84          & 64.38          & 59.62          & 54.85          & 39.40     & 17.84     & 31.79          & 28.21          & 20.60          \\
                      & DI           & 78.18          & 99.81          & 61.75          & 60.04          & 56.15          & 40.56     & 10.80     & 42.76          & 42.01          & 34.28          \\
                       & SGM          & 86.60          & 99.67          & 72.20          & 62.34          & 56.36          & 45.42     & 17.66     & 41.45          & 37.85          & 29.41          \\
                       & IAA          & 93.82          & \textbf{99.78} & 87.98          & 88.26          & 87.02          & 79.12   & /       & 61.02          & 53.80          & 46.34          \\
                       & \pas & \textbf{96.06} & 99.76          & \textbf{90.94} & \textbf{91.00} & \textbf{88.12} & \textbf{85.96} & \textbf{51.74} & \textbf{75.08} & \textbf{72.22} & \textbf{62.28} \\
\bottomrule
\end{tabular}
}
\caption{
Attack success rate (\%) against normally trained and defense models on \subsetfive{} compared to classical, ensemble-based, and structure-based attackers.
The best results are in \textbf{bold}. / indicates the lack of results.
}
\label{tab:5000-all}
\end{table*}

\section{Experiments}

In this section, we conduct extensive experiments to answer the following questions: Is \pas{} effective to craft adversarial examples with high transferability against normally trained (\textbf{RQ1}) and defense (\textbf{RQ2}) models?
How do different parts of \pas{} affect its performance? (\textbf{RQ3})
How does \pas{} affect adversarial transferability? (\textbf{RQ4})

\subsection{Experiment Setup}

\noindent \textbf{Dataset.}
To compare with baselines, we report the results on two datasets:
1) \subsetone{}: ImageNet-compatible dataset in the NIPS 2017 adversarial competition~\cite{kurakin20181000}, which contains 1000 images;
2) \subsetfive{}: a subset of ImageNet validation images, which contains 5000 images and is used by SGM and IAA.
We check that all models are almost approaching the $100\%$ classification success rate.

\noindent \textbf{Models.}
We conduct experiments on both normally trained models and defense models.
For normal trained models, we consider 8 models containing VGG19~\cite{simonyan2014vgg}, ResNet-152 (RN152)~\cite{he2016resnet}, DenseNet-201 (DN201)~\cite{huang2017densenet}, Squeeze-and-Excitation network (SE154)~\cite{hu2018senet}, Inception-v3 (IncV3)~\cite{szegedy2016incv3}, Inception-v4 (IncV4), Inception-Resnet-v2 (IncRes)~\cite{szegedy2017incv4res} and ViT-B/16 (ViT)~\cite{dosovitskiy2020vit}.
%
For defense models, we select 3 robustly trained models using ensemble adversarial training~\cite{tramer2018ens}: the ensemble of 3 IncV3 models (IncV3\ens{3}), the ensemble of 4 IncV3 models (IncV3\ens{4}) and the ensemble of 3 IncResV2 models (IncResV2\ens{3}).
We choose different models (\ie, RN152, DN201, RN50, RN121, IncV4, and IncResV2) as surrogate models to compare with different baselines.
VGG19 is used as the validation model for path evaluation except in \textbf{RQ3}.

\noindent \textbf{Baseline Methods.}
To demonstrate the effectiveness of \pas{}, we compare it with existing competitive methods,
\ie,
classical attackers I-FGSM~\cite{kurakin2018ifgsm}, MI~\cite{dong2018mi}, DI~\cite{xie2019di} and Admix~\cite{Wang2021admix}; structure-based attackers SGM~\cite{wu2019sgm}, LinBP~\cite{guo2020linbp}, IAA~\cite{zhu2021aai}, LLTA~\cite{fang2022llta}; feature-level attackers FDA~\cite{AdityaGaneshan2019fda}, FIA~\cite{ZhiboWang2021fia} and NAA~\cite{zhang2022naa}; and ensemble-based attackers Ens~\cite{LiuCLS17ensi} and SVRE~\cite{xiong2022svre}.
Since part hyperparameters differ between these methods, we directly use their paper results.

\noindent \textbf{Metrics.}
Following the most widely adopted setting, we use the attack success rate as the metric.
Specifically, the Attack Success Rate (ASR) is defined as the percentage of adversarial examples that successfully mislead the victim model among all adversarial examples generated by the attacker.

\noindent \textbf{Hyperparameter.}
For the search process in \pas{}, we conduct $N_{t}=2000$ trials to search on the DAG for each surrogate model, which evaluates the transferability of the backpropagation path on $256$ examples in one-step attacks against the validation model (\ie, VGG19).
For a fair comparison of different datasets with baselines, we use the corresponding baselines' hyperparameter setting.

\noindent \textbf{Extra overhead.}
The extra overhead of \pas{} comes from the search process.
It is approximately 20 times more to generate adversarial samples than a 10-step attack on \subsetone{}.
Note that the overhead of PAS is fixed for a given surrogate model.
When the searched path is used for \subsetfive{} or a larger test set, the relative overhead is linearly reduced.

\begin{table*}[ht]
    \centering
    \scalebox{1.0}{
    \begin{tabular}{@{}llrrr|llrrr@{}}
    \toprule
                           & Attacker & IncV3\ens{3} & IncV3\ens{4} & IncRes\ens{3} &                         & Attacker     & IncV3\ens{3}         & IncV3\ens{4}         & IncRes\ens{3}      \\ \midrule
    \multirow{5}{*}{\rotatebox{90}{RN50}}  & I-FGSM   & 17.3  & 18.5  & 11.2     & \multirow{5}{*}{\rotatebox{90}{DN121}}  & I-FGSM   & 21.8  & 21.5  & 13.1      \\
                           & SGM      & 30.4  & 28.4  & 18.6     &                         & SGM      & 36.8  & 36.8  & 22.5   \\
                           & LinBP    & 34.5  & 32.5  & 20.9     &                         & LinBP    & 39.3  & 38.3  & 22.6           \\
                           & LLTA     & 50.6  & 47.3  & 33.6     &                         & LLTA     & 59.1  & 60.5  & 46.8          \\
                           & \pas{}      & \textbf{72.8} & \textbf{70.4}  & \textbf{57.9}     &                       & \pas{}      & \textbf{70.9}  & \textbf{70.8}  & \textbf{57.4}    \\
    \bottomrule
    \end{tabular}
    }
    \caption{Attack success rate (\%) against robustly trained models on \subsetone{} compared to structure-based attackers. The best results are in \textbf{bold}.}
    \label{tab:all-defense-llta}
\end{table*}

\begin{table*}[ht]
    \centering
    \scalebox{0.9}{
    \begin{tabular}{@{}llrrr|llrrr@{}}
    \toprule
                           & Attacker & IncV3\ens{3} & IncV3\ens{4} & IncRes\ens{3} &                         & Attacker     & IncV3\ens{3}         & IncV3\ens{4}         & IncRes\ens{3}      \\ \midrule
    \multirow{5}{*}{\rotatebox{90}{IncV4}}  & MI-PD     & 23.9          & 24.5          & 12.5       & \multirow{5}{*}{\rotatebox{90}{IncRes}} & MI-PD     & 28.8          & 26.7          & 16.3             \\
                           & FIA-MI-PD & 45.5          & 42.1          & 23.5        &  & FIA-MI-PD & 49.7          & 44.9          & 31.9            \\
                           & NAA-MI-PD & 55.4          & 53.6          & 34.4          & & NAA-MI-PD & 61.9          & 59.0          & 48.3          \\
                           & Admix-MI-DI & 62.4          & 69.3          & 39.7          & & Admix-MI-DI & 70.5          & 63.7          & 55.3          \\
                           & \pas{}-MI-DI    & \textbf{71.5} & \textbf{66.8} & \textbf{49.7} & & \pas{}-MI-DI    & \textbf{76.9} & \textbf{71.2} & \textbf{59.8}  \\
    \bottomrule
    \end{tabular}

    }
    \caption{Attack success rate (\%) against robustly trained models on \subsetone{} compared to classical and feature-level attackers. The best results are in \textbf{bold}.}
    \label{tab:all-defense-feat}
\end{table*}

\subsection{Attack Normally Trained Models (RQ1)}

In this part, we investigate the transferability of attackers against normally trained models on \subsetfive{}.
We report the attack success rates of \pas{}, baselines, ensemble-based and structure-based attackers with RN152 and DN121 as the surrogate model on \subsetfive{} in \cref{tab:5000-all}.

\cref{tab:5000-all} demonstrates that \pas{} beats other attackers in all black-box scenarios.
Averagely, \pas{} achieves 88.13\% ASR for RN152, which is 5.62\% higher than IAA and 20.90\% higher than SGM.
For DN201, \pas{} achieves an average improvement of 2.25\% in comparison with IAA, and we observe a better improvement for \pas{} in victim models, which are more difficult to attack (\eg, 6.84\% improvement against IncRes).
Since SGM manually tunes the decay factors for \skipgrad{} and IAA uses Bayesian Optimization for \skipgrad{} and \skiprelu{}, we owe the improvement to both the DAG search space and the efficient one-step approximation of \pas{}, which boosts adversarial transferability.

As access to the validation model (\ie, VGG19) is permitted, we compare \pas{} with the naive and SOTA ensemble-based attackers.
\cref{tab:5000-all} shows the attack success rates of Ens and SVRE by simply ensemble VGG19 with white-box surrogate models.
The results demonstrate that \pas{} takes full advantage of the additional surrogate models to evaluate the transferability of backpropagation paths, and improves the success rate by a huge margin.

Moreover, the improvement of attack success rates in various victim models (\eg, classical CNNs and ViT, which is a transformer-based vision model), shows the searched path is not overclaimed to the validation model and \pas{} is effective in improving the adversarial transferability.

\subsection{Attack Defense Models (RQ2)}

In this part, to further verify the superiority of \pas{}, we conduct a series of experiments against defense models.
We illustrate the attacking results against competitive baseline methods under various experimental settings. 

\cref{tab:5000-all} shows the ASR on \subsetfive{}.
The advantages of \pas{} are more noticeable against defense models.
The average ASR is 52.85\% and 69.86\% for RN152 and DN201, respectively, which is 16\% more than the second-best IAA.

For the commonly used \subsetone{}, we directly attack defense models since most of the existing attackers have achieved a 90\% attack success rate against normally trained models.
The comparisons between \pas{} and the feature-level and structure-based attackers are presented in \cref{tab:all-defense-llta} and \cref{tab:all-defense-feat}, respectively.

According to the experimental results, highly transferable attacks are crafted against defense models in the average of 23.2\% and 10.9\% by \pas{}.
Although LLTA tunes the data augmentation and backpropagation structure through meta tasks, \pas{} searches the DAG and achieves higher transferability in \cref{tab:all-defense-llta}, which shows the improvement that comes with a larger search space.

We further demonstrate that the adversarial transferability of \pas{} can be exploited in combination with existing methods.
In contrast to the results in LLTA that DI conflicts with LinBP and leads to large performance degradation, we combine \pas{} with DI for transferability gains.
As shown in \cref{tab:all-defense-feat}, when combined with both MI and DI, \pas{} improves the SOTA transferability by a huge margin consistently against robustly trained models by at least 11.5\%.

All in all, the experimental results identify higher adversarial transferability of \pas{} against defense models.
Compared with existing attackers, \pas{} achieves a 6.9\%$\sim$24.3\% improvement in ASR and demonstrates the generality with various surrogate models on two benchmarks.

\begin{table}
\centering
\begin{tabular}{@{}lllll@{}}
\toprule
     &        & Normal & Defense & Total \\ \midrule
& \pas{}          & 90.43  & 66.63   & 83.94 \\ \midrule
& Random         & 33.34  & 20.30   & 29.43 \\ \midrule
\multirow{2}{*}{\rotatebox{90}{DAG}}
& w/ \skipgrad{}   & 57.36  & 23.80   & 48.21 \\
& w/ \skipconv{}   & 76.16  & 38.33   & 65.85 \\
\midrule
\multirow{4}{*}{\rotatebox{90}{Val. Model}}
& RN50          & 87.81  & 60.52  & 80.37 \\
& IncV3          & 91.05  & 63.93   & 83.65 \\
&  RN18          & 93.10  & 66.97   & 85.97 \\
&  DN121          & 94.04  & 72.53   & 88.17 \\
\bottomrule
\end{tabular}
\caption{
The contribution of each part in \pas{} (\ie, search algorithm, DAG, and validation models for path evaluation).
We show the statistics of attack success rate (\%) against all victims.
w/ indicates the search space with the skip module.
}
\label{tab:ablation}
\end{table}

\subsection{Ablation Study (RQ3)}

In this part, we conduct the ablation study to verify the contribution of each part in \pas{} by different search algorithms, removing skip modules in DAG and path evaluation with different validation models.
%

\noindent \textbf{Search algorithm.}
We use a random search as the baseline.
The result shows that randomly sampled paths lead to performance degradation.
It not only validates the effectiveness of the search algorithm but also shows the importance of the paths' design.

\noindent \textbf{DAG space.}
We utilize \pas{} on different search spaces to search for the backpropagation path and observe the ASR.
We compare the commonly used \skipgrad{} with the proposed \skipconv{} and the whole DAG search space
The experimental results are reported in \cref{tab:ablation}.
\skipconv{} leverages the gradient from the critical features and achieves the highest attack success rate among all DAGs with a single skip module.
The most transferable path is searched for by combining all skip modules and achieves at least a 13.02\% improvement compared with the variants.

\noindent \textbf{Path evaluation.}
In \textbf{RQ1}, we show that \pas{} does not overfit the validation model by the improvement of transferability on different structures of the victim model (\eg, ResNet-like models, transformer-like models, and ensemble models).
Further, we investigate the impact of different validation models.
We select RN50, IncV3, RN18, and DN201 as validation models in \cref{tab:ablation} for path evaluation.
It is surprising that even when using the surrogate model itself (\ie, RN50) for evaluation, the searched path is not fully overfitted.
Furthermore, despite the similar structure of RN18, the second-ranking is still achieved.
According to the results, we conclude that using a one-step approximation for path evaluation plays the role of regularization that reduces the overfitting of the searched path to the validation model.

\subsection{Adversarial Transferability with PAS (RQ4)}
\label{sec:howork}

As network architecture shifts from manual to automated design, \pas{} attempts to directly use a validation model as the approximation of adversarial transferability and search highly transferable paths in backpropagation.
In this part, Based on the paths searched on different architectures, we explain how PAS affects adversarial transferability and provide more insights for transfer-based attackers.

\begin{figure}[ht]
\centering
\includegraphics[width=0.95\linewidth]{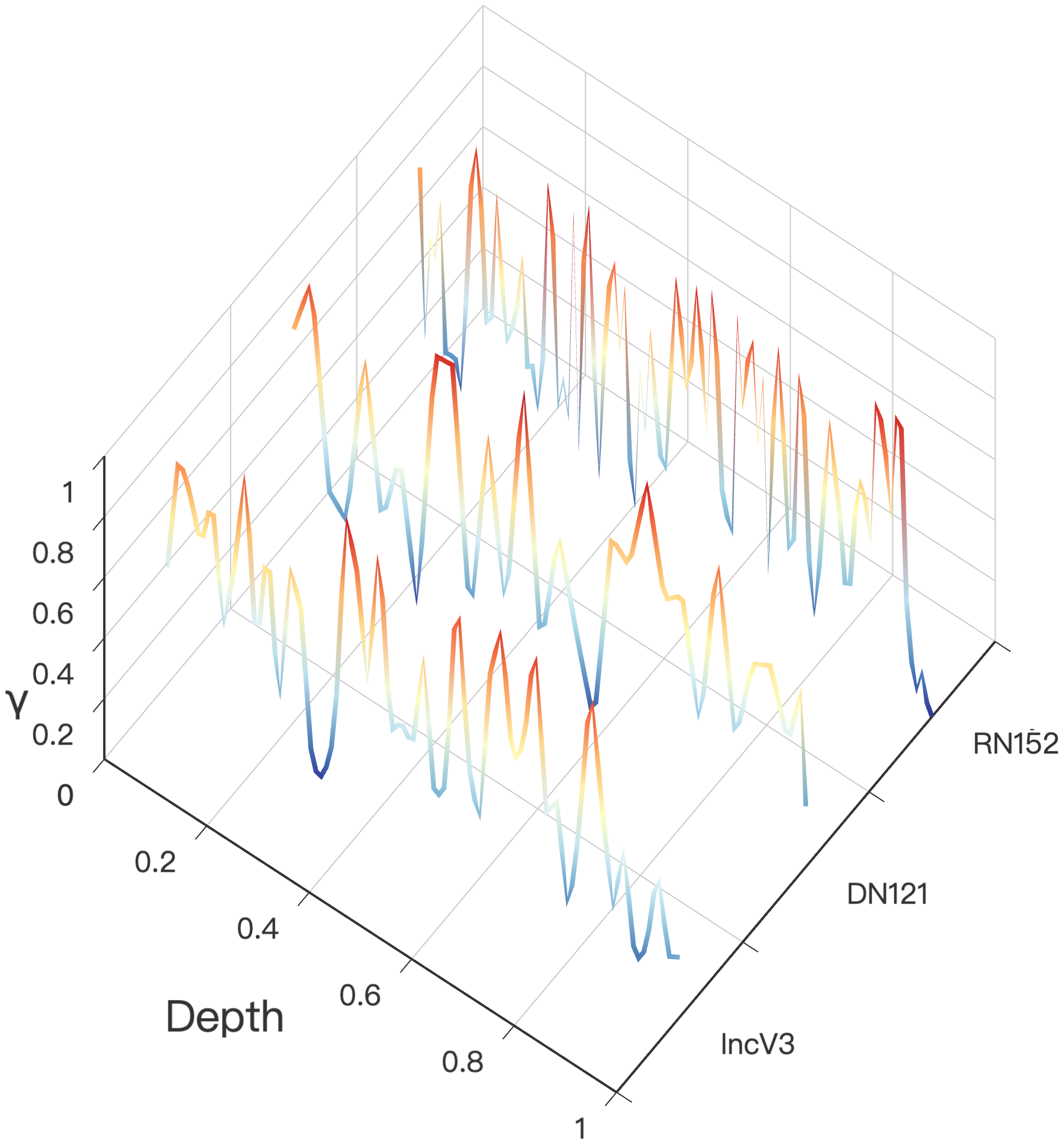}
\caption{
Examples of gradient weights $\Gamma=\{\gamma_i\}$ w.r.t its depth.
The depth indicates the module's discrete index, which is then scaled to a continuous value $\in [0, 1]$.
The final modules maintain a much smaller decay weight, and the intermediate gradients are selectively retrained.
}
\label{fig:gamma-depth}
\end{figure}

\noindent \textbf{Feature attribution.}
\cref{fig:ig-pas} shows the heat map of each input feature importance attributed by integrated gradients on the normal and PAS-reparameterized surrogate models.
We observe that PAS explicitly enhances the attribution of the object for classification and excludes the influence of irrelevant background.
Intuitively, it is more transferable to perturb the object.
Therefore, PAS makes the surrogate model focus on class-related objects by DAG searching to boost adversarial transferability.

\noindent \textbf{Critical feature.}
PAS improves adversarial transferability by selecting critical features through the weighted gradient of features in DAG.
%
%
We plot the distribution of gradient weights $\Gamma = \{ \gamma_i \}$ in \cref{fig:gamma-depth} and find that the final layers often keep a much smaller $\gamma_i$, \ie, skipping their gradients is effective.
The decay weights of intermediate modules are irregular, which means that gradients are selectively retained.
To avoid significant loss of information, it is absent that gradients are all skipped in several neighboring layers (\ie, smaller $\gamma$).
The above conclusion is consistent with feature-level transfer-based attacks contaminating specific intermediate features.

\section{Related Work}

Black-box attackers can be roughly divided into query-based and transfer-based schemes.
Query-based attackers estimate gradient with queries of the prediction to the victim model~\cite{papernot2017querypractical,su2019queryone}.
Due to the lack of access to numerous queries in reality, part of query-based attackers focus on improving efficiency and reducing queries.
In contrast, transfer-based attackers do not require any query and can be applied to unaccessible victim models.

To boost adversarial transferability, various methods have been proposed to reduce the overfitting of the attack on the surrogate model.
Regarding adversarial example generation as an optimization process, \cite{dong2018mi,lin2019nisi} leverage momentum terms to escape from poor local optima.
To avoid overfitting with the surrogate model and specific data pattern, data augmentation~\cite{xie2019di,DongPSZ19ti,zhang2023atk2} and model augmentation \cite{LiuCLS17ensi,xiong2022svre} are effective strategies.
Since the most critical features are shared among different DNNs, feature-level attackers \cite{AdityaGaneshan2019fda,zhang2022naa} destroy the intermediate feature maps.

From the backpropagation perspective, structure-based attackers directly rectify the backpropagation path and leverage more gradients from more useful modules to avoid overfitting.
SGM~\cite{wu2019sgm} reduces the gradient from deep residual modules via skip connection.
LinBP~\cite{guo2020linbp} and ConBP~\cite{zhang2021conbp} use more linear and smoother gradients to replace the sparse gradients of the ReLU module.

The above attackers are designed in a heuristic manner.
To automate the parts that require expert solutions, the idea of AutoML \cite{li2017hyperband,darts} can be applied to adversarial attack.
For a particular domain, AutoML summarizes a large search space of parameters and configurations (e.g., DAG) based on expert experience and searches for the optimal solutions using methods such as black-box optimization.
During the search process, a certain metric is required to evaluate each solution.
\cite{fang2022llta,li2020ghost} enhances adversarial transferability through automatic search in the designed space.
However, the restricted search space leads to limited transferability.
Focusing on this branch, we first expand the backpropagation path of convolution modules inspired by RepVGG~\cite{ding2021repvgg} and form the backpropagation DAG.
Then, we propose the unified framework to efficiently search in the DAG.
While NAS (\ie, the common AutoML application) searches the forward architecture of mixture operations, PAS reparameterizes the surrogate's backward path with skip connections and finds the transferable path.

\section{Conclusion}

In this paper, we focus on structure-based attackers and propose \pas{} to search backpropagation paths for adversarial transferability.
To adjust the backpropagation path of convolution, we propose \skipconv{}, which calculates forward as usual but backpropagates loss in a skip connection form through structural reparameterization.
Then, we construct a DAG-based search space by combining the backpropagation paths of various modules.
To search for the optimal path, we employ Bayesian Optimization and further reduce the search overhead by a one-step approximation for path evaluation.
The results of comprehensive attack experiments in a wide range of transfer settings show that \pas{} considerably improves the attack success rate for both normally trained and defense models.
We explore structural reparameterization for the Transformer’s modules,  use PAS to improve adversarial transferability between CNNs and Transformers, and find better paths by advanced search algorithms and search objectives in future work.

{\small
\bibliographystyle{ieee_fullname}
\bibliography{egbib}
}

\clearpage

\appendix

\section{Effectiveness of One-Step Approximation}

In this section, we verify the effectiveness of the one-step approximation through empirical experiments and statistical measures.
Note that, we aim to rank two paths rather than to accurately evaluate their transferability.

\subsection{Number of instances for evaluation $N$.}

\begin{figure}[ht]
\centering
\includegraphics[width=0.48\textwidth]{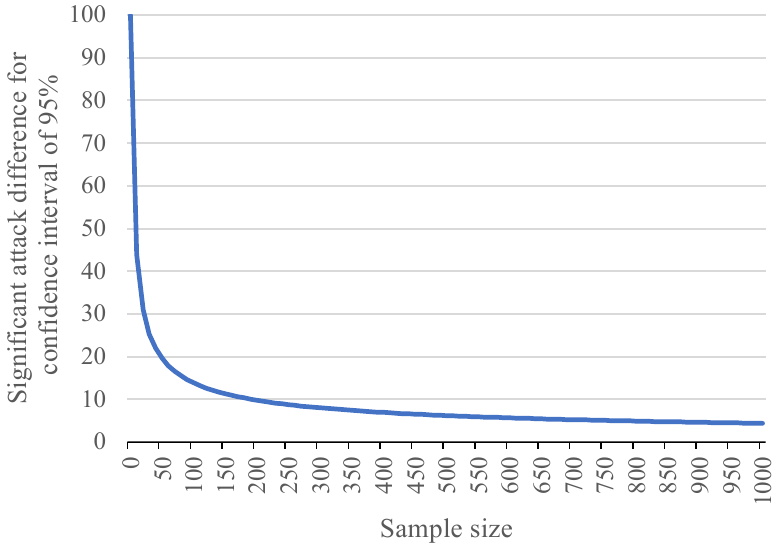}
\caption{Significant difference of attack success rate in the worst case.}
\label{fig:diff-th}
\end{figure}

\begin{figure}[ht]
\centering
\includegraphics[width=0.48\textwidth]{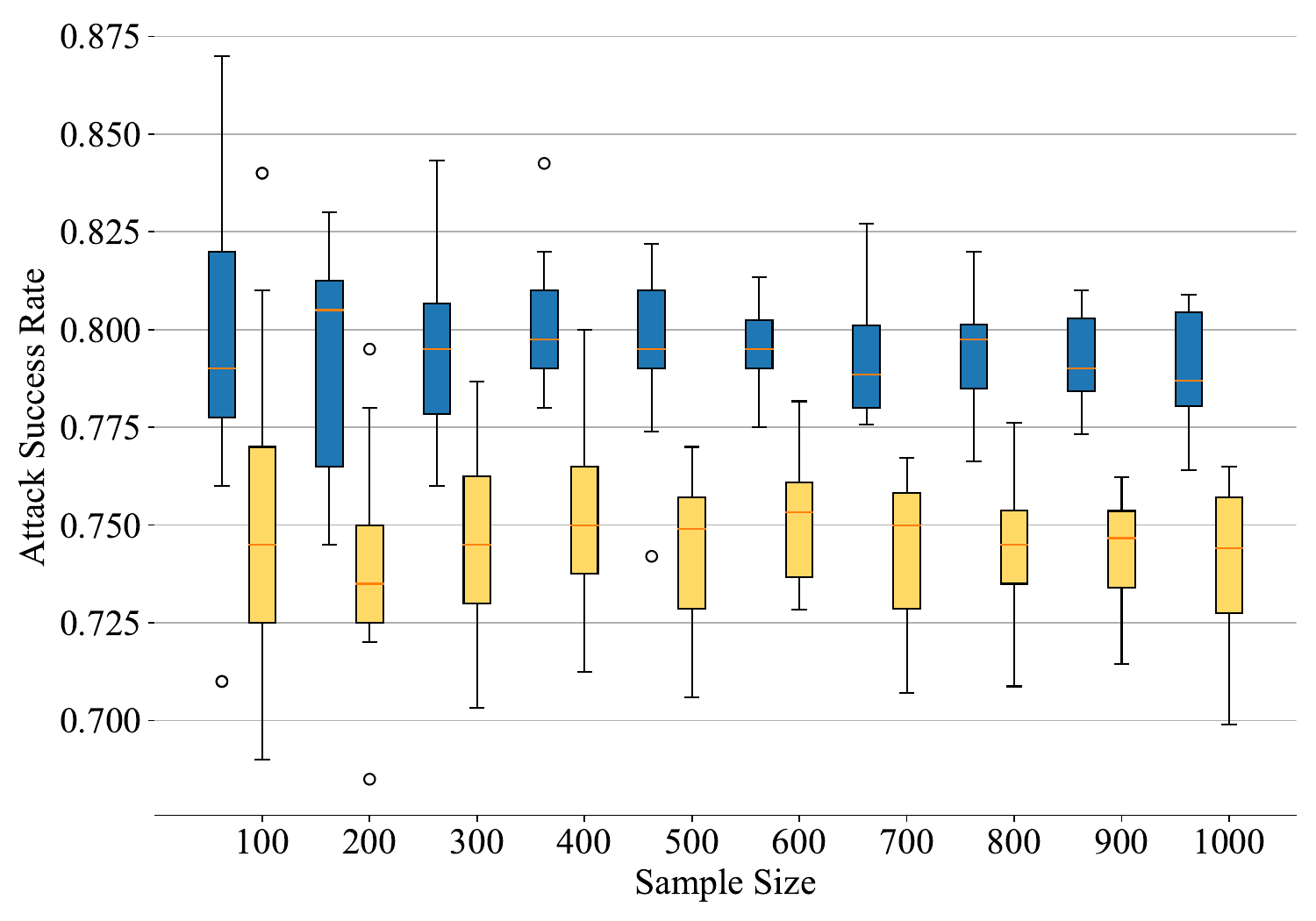}
\caption{Statistics of attack success rates at different sample sizes for paths with $\lvert s_1 - s_2 \rvert = 5\%$ on the test set.}
\label{fig:diff-ss}
\end{figure}

We measure the approximation of transferability by altering the sample size for path evaluation.
We leverage the chi-square test to judge whether there is a significant difference between paths.
Specifically, the question is whether there is a significant difference in the transferability of paths $p_1$ and $p_2$ on a validation dataset of $N$ samples, for which the attack success rate on the test dataset is $s_1$ and $s_2$, respectively.
We calculate the statistic as $stats=\frac{2N(s_1-s_2)^2}{ (s_1+s_2)(2 - s_1 - s_2)}$, which follows $\chi^{2}$ distribution with $1$ degree of freedom.
The corresponding statistics should satisfy $stats \geq 3.841$ for the confidence interval of 95\%.
Thus, we maintain a sliding window to store paths that are not significantly different from the candidates in the search process based on statistical confidence.
Considering the worst case (\ie, $s_1+s_2=1$), we plot the relationship between the sample size $N$ and the significant difference of attack success rate $\lvert s_1 - s_2 \rvert$.
As shown in \cref{fig:diff-th}, we need $200$ samples (\ie, $20\%$ of the test set) to observe better paths with a $10\%$ difference in the attack success rate.
To further verify its effectiveness, we select two paths with a difference of $5\%$ in the test set, sample validation datasets of different sizes 20 times, and draw a box plot.
\cref{fig:diff-ss} shows that the difference in transferability is more obvious as the size of the validation dataset increases.
Moreover, the evaluation of adversarial transferability on 200$\sim$300 samples is sufficient to distinguish between the two paths, except for a few outliers.
Thus, we choose $N=200$ to compare different paths in \pas{}.

\subsection{Number of steps for evaluation.}

\begin{figure}[ht]
\centering
\includegraphics[width=0.48\textwidth]{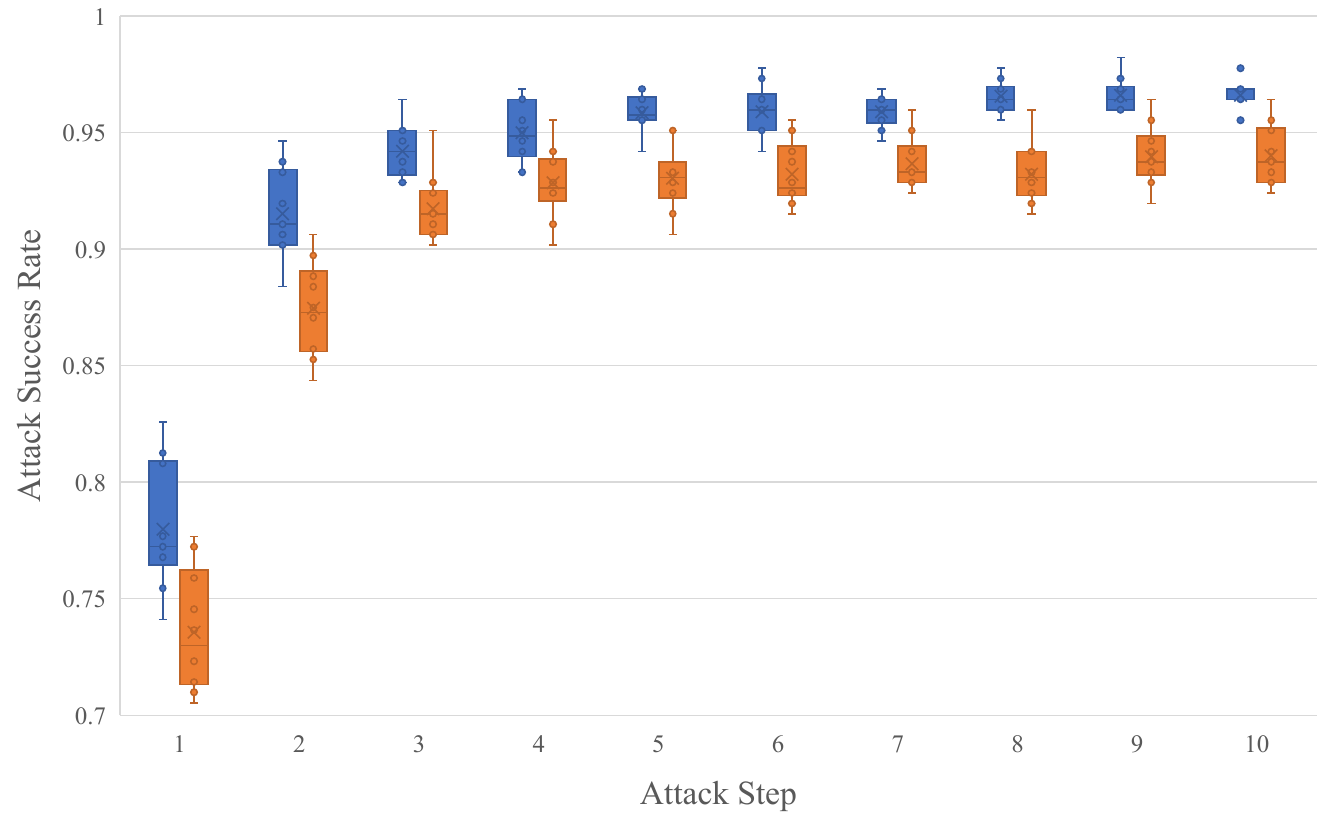}
\caption{Statistics of attack success rates at different attack steps for paths with $\lvert s_1 - s_2 \rvert = 5\%$ on the test set.}
\label{fig:diff-step}
\end{figure}

We measure the impact of the number of attack steps on path evaluation with $N=200$.
Following the experimental settings mentioned above, we sample two paths with a significant difference (\ie, 5\%) and calculated the attack success rate from 1 to 10 attack steps in the validation set.
\cref{fig:diff-step} shows that as the attack steps increase, the attack success rate rises with a diminishing marginal effect.
At different step numbers, the ASR of the paths still maintains certain differences.
Therefore, we can evaluate the paths with a one-step approximation, and choose the more transferable path accordingly.

\section{Integrated Gradients for \pas{}}

\begin{figure}[hb]
\vspace{-5em}
    \begin{subfigure}{1.0\linewidth}
    \includegraphics[width=1.0\linewidth]{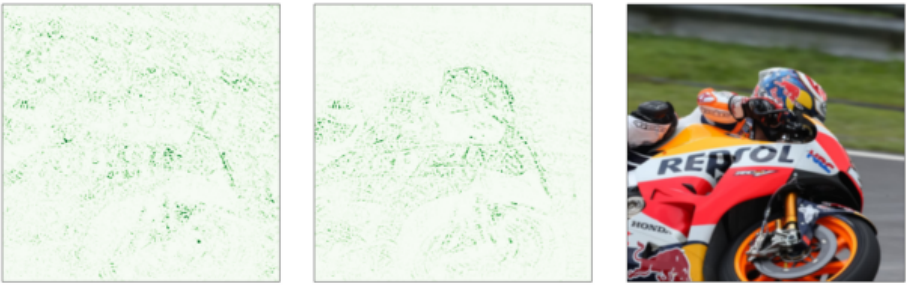}
    \end{subfigure}
    \begin{subfigure}{1.0\linewidth}
    \includegraphics[width=1.0\linewidth]{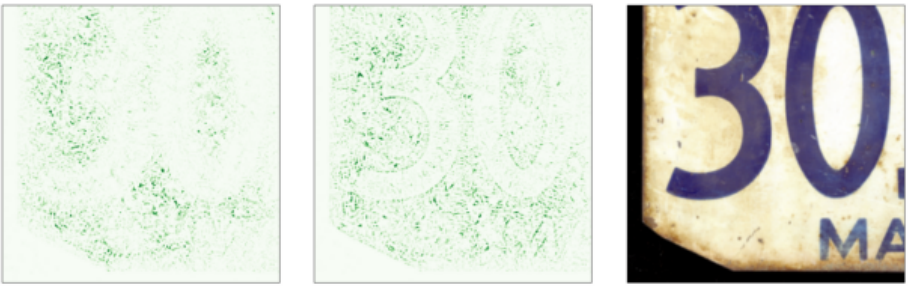}
    \end{subfigure}
    \begin{subfigure}{1.0\linewidth}
    \includegraphics[width=1.0\linewidth]{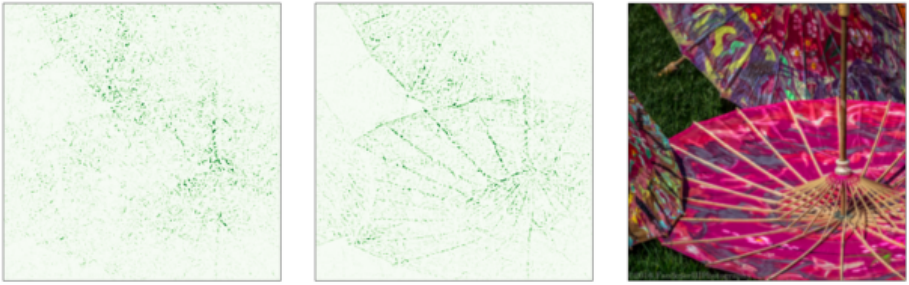}
    \end{subfigure}
    \begin{subfigure}{1.0\linewidth}
    \includegraphics[width=1.0\linewidth]{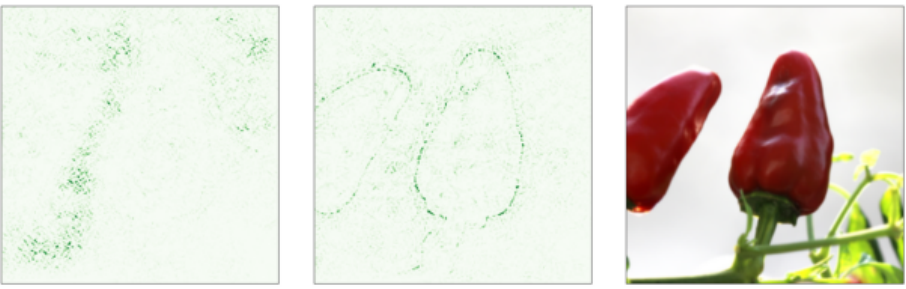}
    \end{subfigure}
    \begin{subfigure}{1.0\linewidth}
    \includegraphics[width=1.0\linewidth]{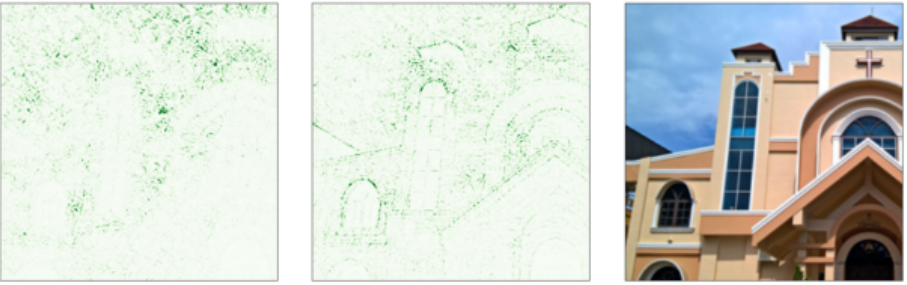}
    \end{subfigure}
    \begin{subfigure}{1.0\linewidth}
    \includegraphics[width=1.0\linewidth]{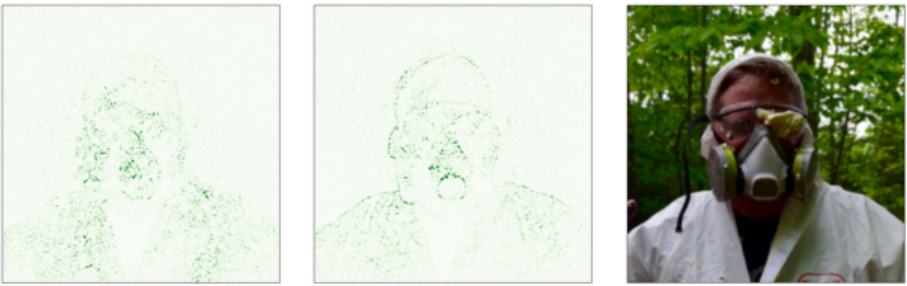}
    \end{subfigure}
    \setcounter{subfigure}{0}
    \begin{subfigure}{0.32\linewidth}
    \centering
    \caption{Surrogate}
    \end{subfigure}
    \begin{subfigure}{0.32\linewidth}
    \centering
    \caption{PAS Surrogate}
    \end{subfigure}
    \begin{subfigure}{0.32\linewidth}
    \centering
    \caption{Original Image}
    \end{subfigure}
\label{fig:more-pas-example}
\end{figure}

In this section, we show more heat maps of IG on the normal and PAS-reparameterized surrogate models.
\cref{fig:more-pas-example} demonstrates that PAS explicitly makes the surrogate model focus on class-related objects and excludes the influence of irrelevant background by DAG searching to boost adversarial transferability.
We believe that PAS plays the role of a denoising model, which removes the overfitting information in the surrogate model and leaves only the gradient part consistent with the data distribution respectively.
%
%
More examples can be found in the supplementary material.

\begin{figure}[ht]
    \begin{subfigure}{1.0\linewidth}
    \includegraphics[width=1.0\linewidth]{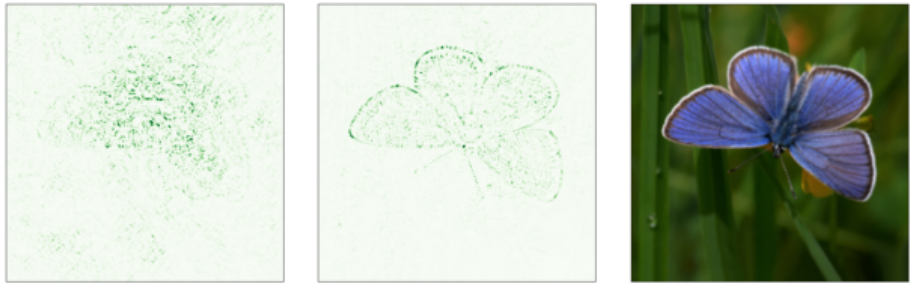}
    \end{subfigure}
    \begin{subfigure}{1.0\linewidth}
    \includegraphics[width=1.0\linewidth]{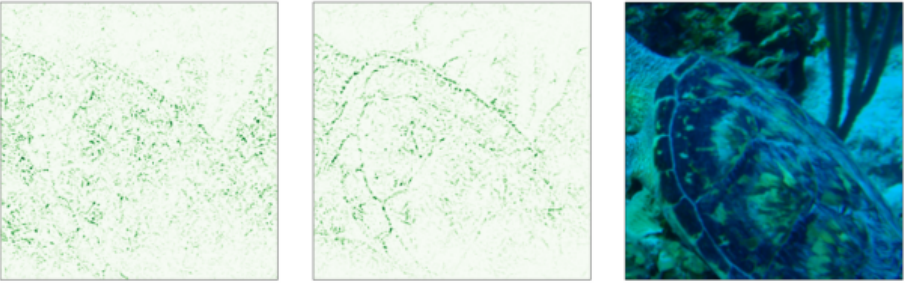}
    \end{subfigure}
    \begin{subfigure}{1.0\linewidth}
    \includegraphics[width=1.0\linewidth]{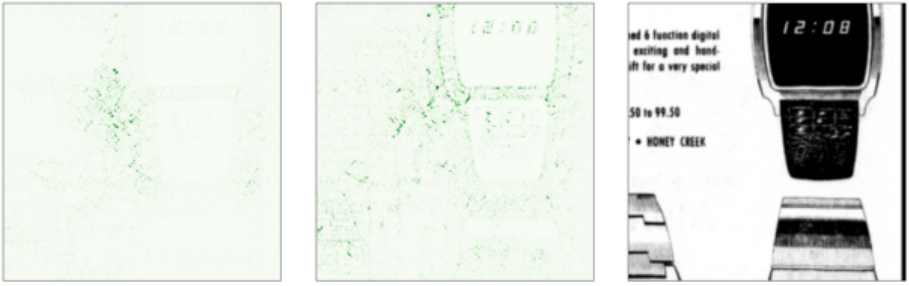}
    \end{subfigure}
    \begin{subfigure}{1.0\linewidth}
    \includegraphics[width=1.0\linewidth]{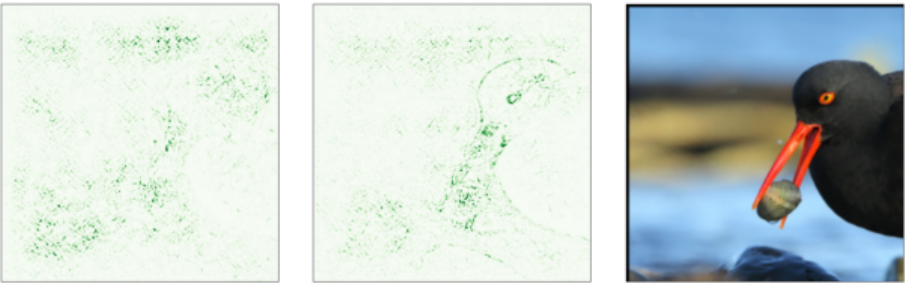}
    \end{subfigure}
    \begin{subfigure}{1.0\linewidth}
    \includegraphics[width=1.0\linewidth]{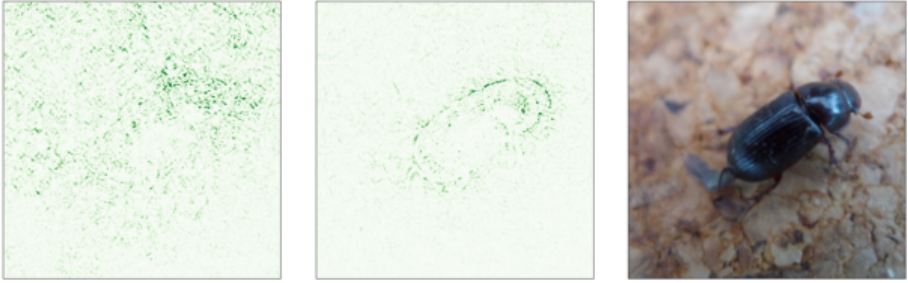}
    \end{subfigure}
    \begin{subfigure}{1.0\linewidth}
    \includegraphics[width=1.0\linewidth]{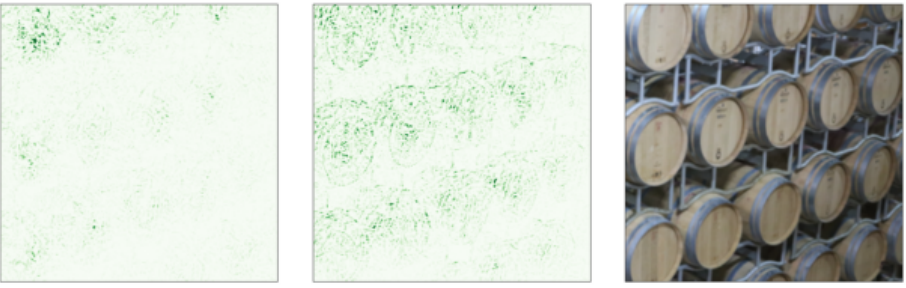}
    \end{subfigure}
    \begin{subfigure}{1.0\linewidth}
    \includegraphics[width=1.0\linewidth]{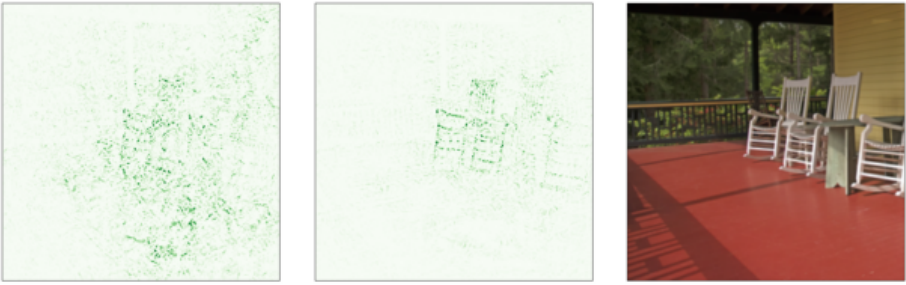}
    \end{subfigure}
    \begin{subfigure}{1.0\linewidth}
    \includegraphics[width=1.0\linewidth]{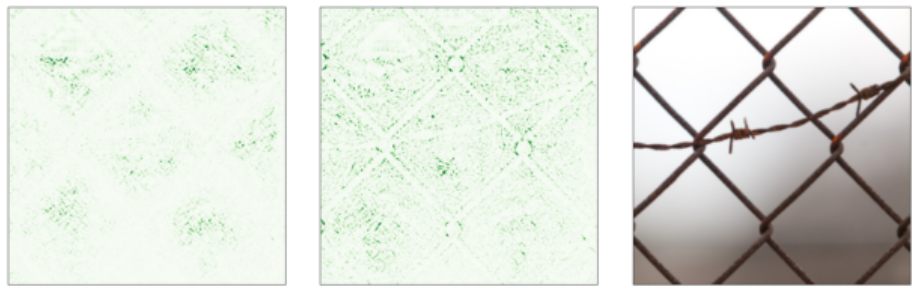}
    \end{subfigure}
    \begin{subfigure}{0.32\linewidth}
    \centering
    \caption{Surrogate}
    \end{subfigure}
    \begin{subfigure}{0.32\linewidth}
    \centering
    \caption{PAS Surrogate}
    \end{subfigure}
    \begin{subfigure}{0.32\linewidth}
    \centering
    \caption{Original Image}
    \end{subfigure}
\end{figure}

\end{document}